\pdfoutput=1
\documentclass[11pt]{article}

\usepackage[preprint]{acl}

\usepackage{times}
\usepackage{latexsym}
\usepackage{tabularx}
\usepackage{multirow}
\usepackage[T1]{fontenc}
\usepackage[utf8]{inputenc}

\usepackage{microtype}

\usepackage{inconsolata}

\usepackage{graphicx}
\usepackage{amssymb}
\usepackage{multicol}
\title{Doc-Guided Sent2Sent++: A Sent2Sent++ Agent with Doc-Guided memory \\for Document-level Machine Translation}

\author{Jiaxin GUO, Yuanchang Luo, Daimeng Wei, Ling Zhang, Zongyao Li, Hengchao Shang\\
    \textbf{Zhiqiang Rao, Shaojun Li, Jinlong Yang, Zhanglin Wu and Hao Yang} \\
        \{jiaxinguo1,luoyuanchang1,weidaimeng,yanghao30\}@huawei.com\\
        Huawei Translation Services Center, Beijing, China }

\begin{document}
\maketitle
\begin{abstract}
The field of artificial intelligence has witnessed significant advancements in natural language processing, largely attributed to the capabilities of Large Language Models (LLMs). These models form the backbone of Agents designed to address long-context dependencies, particularly in Document-level Machine Translation (DocMT). DocMT presents unique challenges, with quality, consistency, and fluency being the key metrics for evaluation. Existing approaches, such as Doc2Doc and Doc2Sent, either omit sentences or compromise fluency. This paper introduces Doc-Guided Sent2Sent++, an Agent that employs an incremental sentence-level forced decoding strategy \textbf{to ensure every sentence is translated while enhancing the fluency of adjacent sentences.} Our Agent leverages a Doc-Guided Memory, focusing solely on the summary and its translation, which we find to be an efficient approach to maintaining consistency. Through extensive testing across multiple languages and domains, we demonstrate that Sent2Sent++ outperforms other methods in terms of quality, consistency, and fluency. The results indicate that, our approach has achieved significant improvements in metrics such as s-COMET, d-COMET, LTCR-$1_f$, and document-level perplexity (d-ppl). The contributions of this paper include a detailed analysis of current DocMT research, the introduction of the Sent2Sent++ decoding method, the Doc-Guided Memory mechanism, and validation of its effectiveness across languages and domains.

\end{abstract}

\section{Introduction}
In the domain of artificial intelligence, Large Language Models (LLMs) like GPT, LLaMa and Qwen play a crucial role in advancements in natural language processing \cite{DBLP:journals/corr/abs-2303-08774,DBLP:journals/corr/abs-2302-13971,DBLP:journals/corr/abs-2307-09288,DBLP:journals/corr/abs-2309-16609,DBLP:journals/corr/abs-2407-10671}. 
LLMs are typically used as a foundation to construct Agents that can unleash their capabilities. In an Agent, Memory \cite{DBLP:conf/coling/LyuD0DWLAWW24} is a key component for addressing long-context dependencies.

Document-level Machine Translation (DocMT) \cite{DBLP:conf/discomt/KimTN19, DBLP:journals/csur/MarufSH21, DBLP:conf/acl/FernandesYNM20} is a very challenging task in the traditional field of machine translation. Thanks to the development of LLMs, DocMT has been receiving growing focus in recent years. \textit{Quality}, \textit{consistency}, and \textit{fluency} are the three core metrics for evaluating DocMT. Recently, some LLMs-based DocMT \cite{DBLP:conf/acl/CuiDZX24, DBLP:conf/wmt/WuH23, DBLP:journals/corr/abs-2401-06468,DBLP:conf/emnlp/WangTWL17, DBLP:conf/emnlp/TanZZ21, DBLP:conf/emnlp/LyuLGZ21} research mainly focuses on constructing DocMT Agents. Typically, these studies use a Memory component to maintain the consistency of text translation across the entire document.

\begin{figure*}[th]
  \centering
  \includegraphics[width=15.8cm]{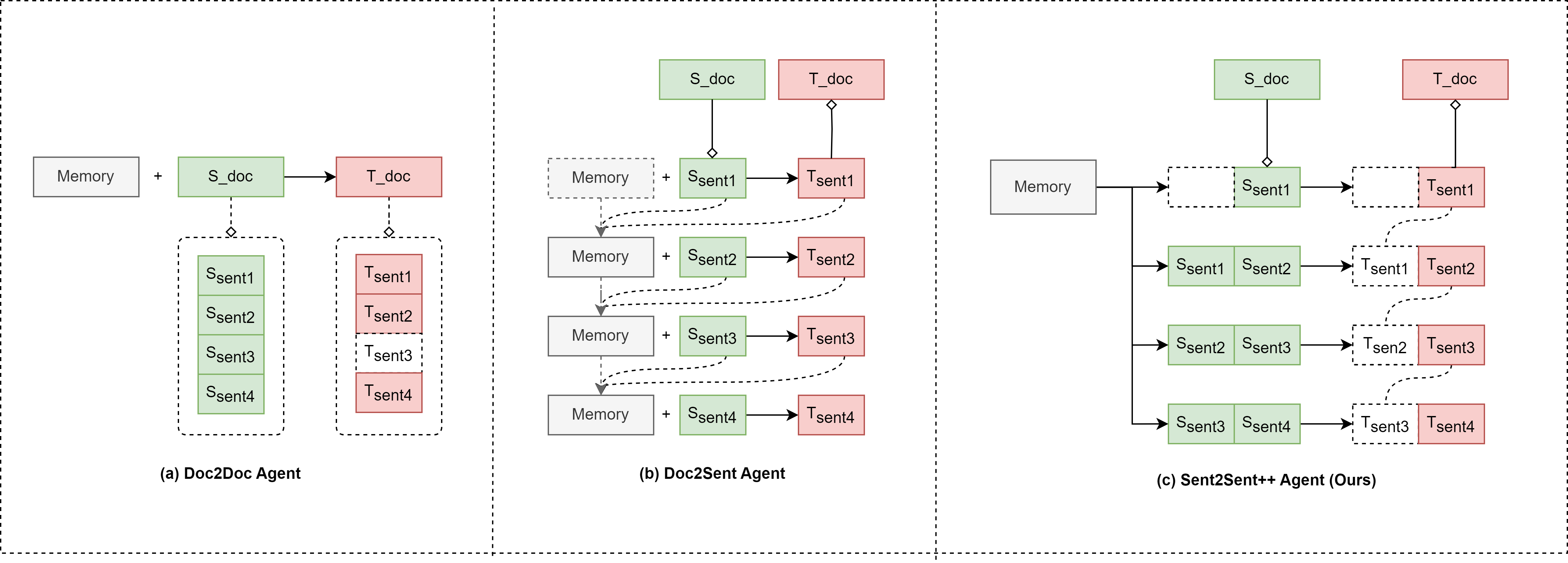}
  \caption{The overview of Current LLM-based DocMT}
  \label{fig:motivation}
\end{figure*}

The decoding strategies of these DocMT Agent research typically fall into two categories: Doc2Doc and Doc2Sent. 
TransAgent \cite{DBLP:conf/wmt/WuH23} addresses the translation of ultra-long texts through Multi-Agent cooperation, often employing a Doc2Doc decoding approach with the aim of achieving smoother translations.However, this method may result in the omission of certain sentences, which can compromise the completeness of the overall translation \cite{DBLP:conf/wmt/KarpinskaI23}. While the omission of individual sentences might be tolerable for novel translations, it is unacceptable for serious literature or news reports, where omissions can be detrimental. 
To tackle the issue of sentence omissions, Delta \cite{DBLP:conf/emnlp/WangTWL17} and IncreD \cite{DBLP:conf/emnlp/LyuLGZ21} have turned to a Doc2Sent decoding framework. These frameworks translate the document by breaking it down into individual sentences, ensuring that each sentence is translated with context information in Memory. Nonetheless, this approach may compromise fluency. 
\textbf{The question is: Is there a strategy that can both avoid sentence omissions and maintain the fluency of the document?}

In this paper, we propose an Agent named Doc-Guided Sent2Sent++, abbreviated as Sent2Sent++, designed to address the challenges of Document-level Machine Translation. Sent2Sent++ employs an incremental sentence-level forced decoding strategy, where each time we decode, we translate two adjacent sentences together, but we only incrementally decode the latter one; the translation of the former sentence is generated in the previous decoding and serves as a prefix in the current decoding process. This way, we ensure that every sentence is translated, and also make the translations of adjacent sentences more fluent, thereby enhancing the fluency of the overall document translation. At the same time, consistent with other research work, our Agent also uses Memory. Specifically, we define this Memory as Doc-Guided Memory, which only contains the summary and its translation of the entire document. We found that, compared to other works using summary, terms, style, and other information in Memory, using only summary information is a very efficient approach. Compared to Doc2Doc and Doc2Sent methods, our approach performs best in terms of quality, consistency, and fluency.

Our main contributions are as follows
\begin{itemize}
    % \item We propose a Doc-Guided Sent2Sent++ Agent, capable of handling document translation tasks and generating high-quality translation results.
    % \item We introduce the Doc-Guided Sent2Sent++ Agent, designed to adeptly manage document translation tasks and consistently produce superior translation outcomes.
    \item We conduct a detailed analysis on current DocMT Agent research, pinpoint the challenges faced, and introduce the Doc-Guided Sent2Sent++ Agent, which adeptly manages document translation tasks and consistently produces superior translation outcomes.
    \item We present a new decoding method Sent2Sent++, which meets the requirements for both sentence completeness and improved document fluency, addressing the limitations of other DocMT Agent approaches.
    \item We introduce a novel Doc-Guided Memory mechanism that leverages abstract bilingual information as key contextual cues, thereby maintaining consistency in document translation.
    \item We validate the effectiveness of our method on test sets across multiple languages, demonstrating its broad applicability across languages and domains.
\end{itemize}

\section{Motivation}

\subsection{Current LLM-based DocMT}

\begin{table*}[ht]
\centering
\begin{tabularx}{\textwidth}{Xccc}
\hline
& Consistency & No Sentence Missing & Fluency \\
\hline
Doc2Doc Agent & $\checkmark$ & $\times$ & $\checkmark$\\
Doc2Sent Agent & $\checkmark$ & $\checkmark$  & $\times$\\
Sent2Sent++ Agent (Ours) & $\checkmark$ & $\checkmark$  & $\checkmark$ \\
\hline
\end{tabularx}
\caption{Which method is more suitable for DocMT? \textbf{Doc2Doc:} high fluency but may lead to sentence-level omissions. \textbf{Doc2Sent: }prioritizes the prevention of sentence-level omissions over translation fluency. \textbf{Sent2Sent++: }ensures the absence of sentence-level omissions while preserving translation fluency.}\label{tab:motivation}
\end{table*}

\paragraph{Doc2Doc Agent}

In the field of DocMT, the Doc2Doc method is a direct approach that translates entire documents from the source language to the target language. As shown in Figure~\ref{fig:motivation}(a), this method maintains the integrity and consistency of the document by incorporating additional information or steps, while also addressing the complex linguistic and cultural nuances within the document. 

Such as TransAgent \cite{DBLP:journals/corr/abs-2405-11804} introduces a multi-agent framework based on LLMs for literary translation, implemented through a virtual translation company named TransAgents. TransAgents emulates the traditional translation and publishing process, harnessing the collective expertise of multiple agents to meet the complex demands of translating literary works. The essence of this framework is the collaboration within the multi-agent system, which includes senior editors, junior editors, translators, localization experts, and proofreaders, all working together to ensure that the translation maintains high quality and consistency throughout the document.

\paragraph{Doc2Sent Agent}

In contrast to the document-level translation approach of Doc2Doc, the Doc2Sent method segments document-level translation tasks into sentence-level tasks. As illustrated in Figure~\ref{fig:motivation}(b), the document is divided into individual sentences, and memory information is integrated during the translation of each sentence to ensure consistency and accuracy throughout the entire document. The choice of memory information extraction methods can lead to varying outcomes.

DelTA \cite{DBLP:journals/corr/abs-2410-08143} is an online document-level translation agent that utilizes a multi-level memory structure to store information of varying granularities and spans, including proper noun records, bilingual abstracts, long-term memory, and short-term memory. These memory components are continuously retrieved and updated by auxiliary LLM-based components. DelTA employs a sentence-by-sentence translation strategy, ensuring no sentence omissions. Additionally, its summarization components has improved the accuracy of pronoun translation.

IncreD \cite{DBLP:conf/wmt/LuoGWSLWRLYY24} proposes a context-sensitive and style-relevant incremental decoding framework specifically tailored for discourse-level literary translation tasks. This framework ensures that the translation of each sentence is informed by its broader context, thereby maintaining the coherence and consistency of the entire text. This approach enables the model to capture long-distance dependencies and stylistic elements, producing translations that faithfully preserve the literary quality of the original text.

\subsection{Which method is more suitable for DocMT?}

As shown in Table~\ref{tab:motivation}, the Doc2Doc Agent capitalizes on the LLM's advanced long-context understanding and generation capabilities to translate entire documents directly from the source to the target language. While this method preserves document consistency and fluency, it occasionally omits paragraphs and sentences. Such omissions are acceptable for entertainment texts, like novels, but can be problematic for serious literary works or news reports. Furthermore, the decoding time is relatively high due to the inclusion of extensive long memory information and lengthy document content. In contrast, the Doc2Sent Agent decomposes document-level translation into sentence-level tasks, using large language models to extract comprehensive information from the document, thus maintaining translation consistency and accuracy. Moreover, this approach ensures no sentence-level omissions. However, aggregating individual sentence translations into a discourse-level translation can diminish sentence fluency. Frequent updates to document memory information and the concatenation of sentence translations result in a considerable increase in the number of calls to large models. 

Therefore, we propose a novel method for LLM-based DocMT – a Doc-Guided Sent2Sent++ Agent. \textbf{This agent not only ensures consistency in document translation but also enhances fluency and guarantees no sentence omissions}, offering superior translation efficiency compared to both Doc2Doc and Doc2Sent Agents.

\begin{figure*}[t]
  \includegraphics[width=15.8cm]{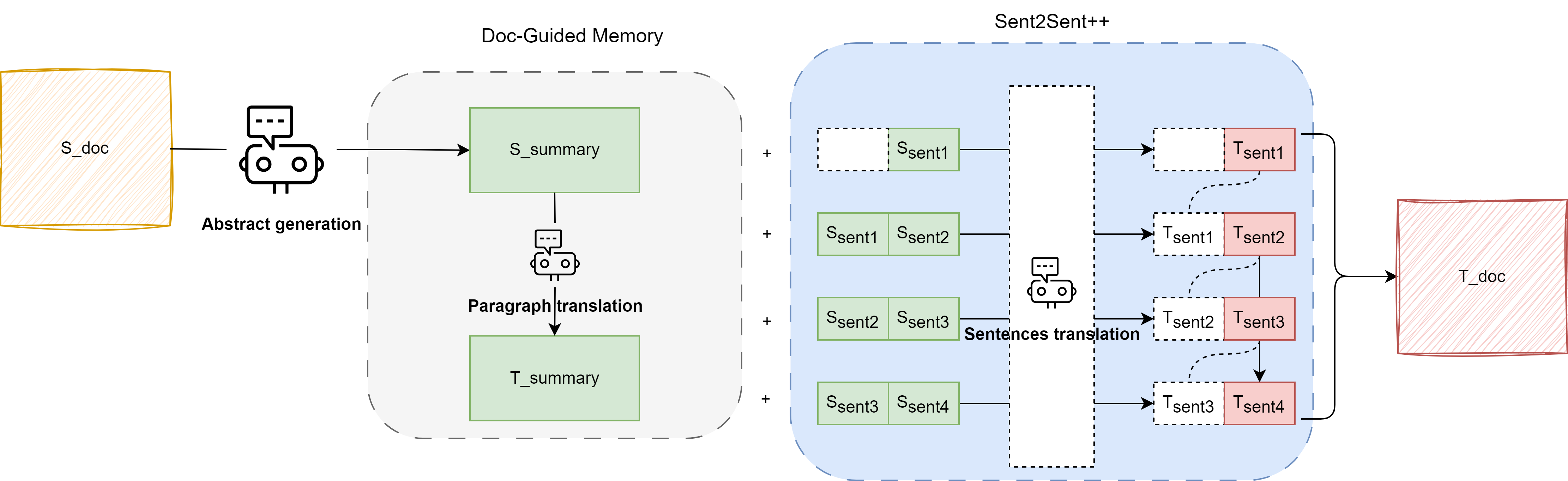}
  \caption{The overall process of Our Doc-Guided Sent2Sent++ Agent}
  \label{fig:Doc-Guided Sent2Sent++ Agent}
\end{figure*}

\section{Approach}

In this paper, we present an Agent titled Doc-Guided Sent2Sent++, or Sent2Sent++ for short, crafted to tackle the complexities of Document-level Machine Translation, see Figure~\ref{fig:Doc-Guided Sent2Sent++ Agent}. Sent2Sent++ utilizes an incremental sentence-level forced decoding strategy, wherein we decode two adjacent sentences in tandem, yet incrementally process only the latter; the prior sentence's translation, generated in the previous round, acts as a prefix in the current decoding phase. This approach not only guarantees the translation of every sentence but also enhances the fluency between adjacent sentences, thereby improving the overall document translation fluency. Moreover, aligning with other studies, our Agent incorporates a Memory component. Notably, we term this Memory as Doc-Guided Memory, which encompasses solely the summary and its translation of the entire document. We've discovered that, in contrast to other studies that employ a variety of information such as summary, terms, and style in Memory, focusing solely on summary information proves to be a highly efficient method. Our method outperforms both Doc2Doc and Doc2Sent approaches in terms of quality, consistency, and fluency.

\subsection{Sent2Sent++}

\begin{figure}[!h]
\centering
  \includegraphics[width=5.0cm]{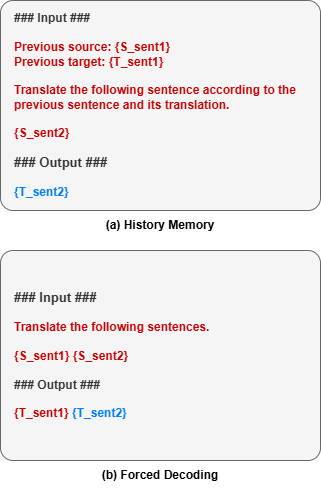}
  \caption{Forced Decoding in Sent2Sent++ Agent}
  \label{fig:Forced Decoding}
\end{figure}

We split the input document into sentences: $S_{doc} = \{S_{sent_1}, S_{sent_2}, S_{sent_3}, ..., S_{sent_n}\}$, Transform document translations to sentence-level translations. Sent2Sent++ forced decoding strategy modifies the input and output processes of sentence-level translation. Specifically, the input is expanded to include two sentences $S_{sent_{i-1}}$ and $S_{sent_{i}}$, with the output of the preceding sentence $T_{sent_{i-1}}$  acting as the mandatory decoding segment for the subsequent process. satisfying the following formula:

\begin{equation}
  \label{Sent2Sent++}
  T_{sent_i} = \mathbb{LLM}(T_{sent_{i-1}}|S_{sent_{i-1}}, S_{sent_i})
\end{equation}

As a result, the final translation, formed by concatenating individual sentence translations: $T_{doc} = (T_{sent_1}, T_{sent_2}, T_{sent_3}, ..., T_{sent_n}) $, inherently exhibits greater fluency. The distinction between the forced decoding strategy in Sent2Sent++ and the storage of context information in History Memory in Doc2Sent is illustrated in Figure \ref{fig:Forced Decoding}.

Furthermore, the integrated Doc-Guided Memory component maintains output consistency, allowing forced decoding to rely on only one sentence to enhance fluency and markedly improve efficiency.

\subsection{Doc-Guided Memory}

Our Doc-Guided Memory is primarily divided into two components. Initially, it leverages the abstractive capabilities of large models to generate a summary of the document: $S_{summary} = \mathbb{LLM}(S_{doc})$, referred to as the summary component. This component encapsulates the general information, terminological details, and stylistic elements of the document. Subsequently, this summary is translated into the target language $T_{summary} = \mathbb{LLM}(S_{summary})$, forming what we call the doc2doc component: $\mathcal{M} = (S_{summary}, T_{summary})$. This bilingual summary continuously guides the sentence-level translation throughout the Sent2Sent++ process. 

Once the Doc-Guided Memory is integrated, the generation process for the sentence $i^{th}$ in the translated document can be articulated as follows:

\begin{equation}
  \label{Doc-Guided Sent2Sent++}
  T_{sent_i} = \mathbb{LLM}(T_{sent_{i-1}}|S_{sent_{i-1}}, S_{sent_i}, \mathcal{M})
\end{equation}

The summary component generates an abstract that ensures the consistency of discourse translation. Unlike \cite{DBLP:journals/corr/abs-2410-08143}, we generate the Doc-Guided Memory based on the original document at the outset, without the need for subsequent updates, thereby enhancing translation efficiency.

The bilingual abstracts generated by the doc2doc component serve as a guide for subsequent translations, ensuring terminological consistency and stylistic coherence throughout the process. As this information acts as a prefix guidance, there is no concern regarding the omission of generated summaries or the possibility that their translation might be overlooked.

\section{Experimental Setup}

\subsection{Datasets}

We conducted experiments on two distinct test sets. The first is the test set from the IWSLT2017 translation task 6\footnote{https://wit3.fbk.eu/2017-01-d/} \cite{DBLP:conf/iwslt/CettoloFBNSSYF17}, comprising parallel documents from TED talks. Our experiments encompassed eight language pairs: English-German, English-French, English-Japanese, and English-Chinese in both directions. Each language pair featured 10 to 12 sentence-aligned parallel documents, with approximately 1,500 sentences per language pair. The second test set is GuofengWebnovel7\footnote{https://github.com/longyuewangdcu/GuoFeng-Webnovel/} \cite{DBLP:conf/wmt/LuoGWSLWRLYY24, DBLP:conf/wmt/WangLLJWXTGCWZK24}, a high-quality, discourse-level web novel corpus. On this corpus, we performed English-Chinese bidirectional experiments using the 'GuoFengV1TEST2' set. Utilizing these two test sets, we have validated the effectiveness of the proposed agent across a range of languages, encompassing eight distinct language pairs, as well as across various domains, including TED talks and novels.

\subsection{Models}

In this paper, we employed two open-source large language models: Qwen-2.5-7B-Instruct\footnote{https://huggingface.co/Qwen/Qwen2.5-7B-Instruct} for generating Chinese translations and Llama-3.1-8B-Instruct\footnote{https://huggingface.co/meta-llama/Llama-3.1-8B-Instruct} for generating other translations. To ensure more stable results, we set the temperature coefficient to 0 and employed beam decoding with a beam width of 5. When utilizing these large models directly for document translation, we set the maximum number of new tokens to 16,384.

\subsection{Metrics}

In the domain of document translation, quality, consistency, and fluency are the three pivotal criteria for assessing the performance of translation systems. Consequently, this paper introduces automatic evaluation metrics focused on these three core criteria. 

\paragraph{Quality Metric} Quality primarily indicates the accuracy of the translated text, hence we assess it using established machine translation evaluation metrics such as s-COMET, d-Bleu and d-COMET \cite{DBLP:conf/wmt/VernikosTMF22}. We utilize the model Unbabel/wmt22-comet-da\footnote{https://huggingface.co/Unbabel/wmt22-comet-da} to obtain the s-COMET scores and Unbabel/wmt21-comet-mqm\footnote{https://github.com/amazon-science/doc-mt-metrics/} obtain the d-COMET scores. 

\paragraph{Consistency Metric} Consistency primarily indicates the terminological coherence and completeness in document translation, thus we evaluate it using LTCR-$1_f$ \cite{DBLP:journals/corr/abs-2410-08143}. 

\paragraph{Fluency Metric}Fluency, which denotes the naturalness and smoothness of the generated translation text, is assessed by calculating the perplexity (ppl) of the translation with the Llama-3.1-8B\footnote{https://huggingface.co/meta-llama/Llama-3.1-8B}.

\begin{table*}[ht]
\centering
\begin{tabularx}{\textwidth}{c|c|X|ccccc}
\hline
\textbf{domain} & \textbf{lang-pair} & \textbf{method} & \textbf{s-comet} & \textbf{d-bleu}  & \textbf{d-comet} & \textbf{LCTR-$1_f$}  & \textbf{d-ppl} \\
\hline
\multirow{12}*{IWSLT17} & \multirow{6}*{zh2en} & Sentence & 83.22 & 26.6 & 7.14 & 87.69 & 11.28 \\
 &  & Doc2Doc & -  & 23.25 & - &  93.05 & 9.19 \\
 &  & Doc2Sent & 83.34  & 26.8 & 7.15 &  92.84  & 11.1 \\
 &  & \ \ \ \ -\ Memory & 83.27  & 26.3 & 7.12 &  87.96  & 11.3 \\
\cline{3-8}
 &  & Sent2Sent++ & 83.34  & 28.3 & 7.25 &  92.85  & 10.49 \\
 &  & \ \ \ \ -\ Memory & 83.39  & 27.9 & 7.27 &  87.47  & 10.32 \\
\cline{2-8}
 & \multirow{6}*{en2zh} & Sentence & 82.9  & 35.1 & 6.41 &  83.79  & 18.25 \\
 &  & Doc2Doc & -  & 36.58 & - &  90.65 & 16.69 \\
 &  & Doc2Sent & 83.42  & 35.7 & 6.44 & 92.14 & 17.53 \\
 &  & \ \ \ \ -\ Memory & 83.26  & 35.3 & 6.43 & 83.93 & 17.54 \\
\cline{3-8}
 &  & Sent2Sent++ & 84.11  & 35.3 & 6.66 & 92.12  & 16.18 \\
 &  & \ \ \ \ -\ Memory & 83.94  & 34.9 & 6.56 & 83.95 & 16.23 \\
\hline
\hline
\multirow{12}*{GUOFENG} &  \multirow{6}*{zh2en} & Sentence & 74.38  & 15.6 & 2.95 & 55.63  & 11.25 \\
 &  & Doc2Doc & -  & 12.76 & - &  84.61 &  9.03 \\
 &  & Doc2Sent & 75.5 & 15.7 & 2.98 & 93.79 & 11.18 \\
 &  & \ \ \ \ -\ Memory & 75.21 & 15.2 & 2.93 & 70.24 & 11.09 \\
\cline{3-8}
 &  & Sent2Sent++ & 76.31  & 16.3 & 3.14 & 94.81  & 10.16 \\
 &  & \ \ \ \ -\ Memory & 76.22  & 16.1 & 3.12 & 68.95  & 10.24 \\
\cline{2-8}
 & \multirow{6}*{en2zh} & Sentence & 80.53  & 25.4 & 4.5 & 50  & 20.43 \\
 &  & Doc2Doc & -  & 28.09 & - &  89.64 & 15.35 \\
 &  & Doc2Sent & 80.62  & 25.8 & 4.49 & 92.57 & 19.76 \\
 &  & \ \ \ \ -\ Memory & 80.23  & 25.1 & 4.46 & 62.13 & 19.74 \\
\cline{3-8}
 &  & Sent2Sent++ & 81.45  & 26.3 & 4.62 & 94.62  & 18.44 \\
 &  & \ \ \ \ -\ Memory & 81.21  & 25.6 & 4.51 & 63.85  & 18.51 \\
\hline
\end{tabularx}
\caption{\centering The English-Chinese translation results for the IWSLT2017 and Guofeng test sets}\label{tab:Results_two_test}
\end{table*}

\subsection{Methods}

We include the following approaches as our Methods. It is important to note that all methods utilize the same LLMs and the same parameter settings.

\begin{itemize}
    \item \textbf{Sentence}: We utilize LLMs for a sentence-level translation process to establish baseline results. 
    \item \textbf{Doc2Doc}: Employing a Large Language Model (LLM), text translation results are derived directly from the original document. Since the test data consists of paragraphs rather than complete texts, using the LLM for paragraph translation already yields high consistency, hence, we have not specifically added a Memory component here.
    \item \textbf{Doc2Sent}: The document is segmented into sentence-level units, information is appended to each sentence, and a Large Language Model (LLM) is used to translate these units, followed by concatenating the text translation results. For a fair comparison, the Doc2Sent Agent employs the Doc-Guided Memory used in this method.
    \item \textbf{Sent2Sent++ (Ours)}: This method extracts document-level information, divides the document into sentence-level segments, incorporates Doc-Guided information into each sentence, and employs an LLM to forcibly decode the previous sentence, generate the current sentence translation, and concatenate the text translation results.
\end{itemize}

\section{Results}
All results indicate that compared to the Doc2Doc method, our approach has fewer omissions and a higher d-BLEU score; compared to the Doc2Sent method, our approach offers better fluency and a lower d-ppl value.

\subsection{The results for the Multi-domain test sets} 

Table~\ref{tab:Results_two_test} presents the results of Chinese-to-English and English-to-Chinese document translation on the IWSLT17 and GUOFENG test sets. To assess the quality of the document translation, we utilized five automatic evaluation metrics.

In terms of consistency, the Sentence method exhibits a notable deficit compared to the other three methods, with a difference of 4-5 in LTCR-$1_f$ values. This discrepancy is expected, as independent sentence translation in discourse naturally results in contextual and terminological inconsistencies. The LTCR-$1_f$ values for the Doc2Sent method and the Doc-Guided Sent2Sent++ method closely align with those of the Doc2Doc method, When memory information is removed, the LTCR-$1_f$ values decrease significantly. it shows that incorporating discourse information during sentence-level translation with LLMs can effectively maintain discourse consistency.

\begin{table*}[th]
\centering
% \footnotesize
\begin{tabularx}{\textwidth}{c|X|ccccc}
\hline
\textbf{lang-pair} & \textbf{method} & \textbf{s-comet} & \textbf{d-bleu}  & \textbf{d-comet} & \textbf{LCTR-$1_f$} & \textbf{d-ppl} \\
\hline
\multirow{6}*{xx2en} & Sentence & 84.25
 & 30.24 & 6.96 & 81.09 & 11.66 \\
 & Doc2Doc & -  & 26.76 & - &  87.87 & 8.8 \\
 & Doc2Sent & 84.39  & 30.23 & 6.99 &  88.78  & 11.36 \\
 & \ \ \ \ \ \ \ -\ Memory & 84.31  & 30.01 & 6.97 &  82.07  & 11.47 \\
\cline{2-7}
 & Sent2Sent++ & 84.4  & 31.49 & 7.08 &  89.31  & 10.69 \\
 & \ \ \ \ \ \ \ -\ Memory & 84.22  & 31.24 & 7.07 &  82.18  & 10.69 \\
\hline
% \cline{2-8}
\multirow{6}*{en2xx} & Sentence & 83.47  & 29.61 & 6.4 &  73.3  & 10.92 \\
 & Doc2Doc & -  & 31.8 & - &  78.83  & 9.53 \\
 & Doc2Sent & 83.72  & 29.88 & 6.44 & 83.63 & 10.65 \\
 & \ \ \ \ \ \ \ -\ Memory & 83.63 & 29.42 & 6.39 & 75.59 & 10.7 \\
\cline{2-7}
 & Sent2Sent++ & 83.77 & 29.83 & 6.55 & 84.36 & 9.69 \\
 & \ \ \ \ \ \ \ -\ Memory & 83.62 & 29.48 & 6.47 & 75.75  & 9.74 \\
\hline
\end{tabularx}
\caption{\centering Results for eight linguistic directions in the IWSLT2017 test set}\label{tab:Results_mul_lang}
\end{table*}

While ensuring discourse consistency, the Doc2Sent method, fundamentally a sentence-level translation method, exhibits a considerably d-ppl value compared to the Doc2Doc method, aligning with the level of the Sentence method. This represents a significant limitation of the Doc2Sent method, specifically its suboptimal sentence fluency. Conversely, the proposed Doc-Guided Sent2Sent++ method, which incorporates the Sent2Sent++ forced decoding method for sentence-level translation, attains a d-ppl value on par with the Doc2Doc method. This demonstrates that the method can indeed yield translations with enhanced fluency.

Although the Doc2Sent method demonstrates satisfactory consistency and fluency in the translation of entire texts, its sentence-level approach inevitably leads to sentence-level omissions, a limitation that is particularly pronounced when the large model's capacity to decode long texts is subpar. This is evident in the d-BLEU metric, with a marked disparity observed in the Chinese-to-English direction. In contrast, the Doc-Guided Sent2Sent++ method employs sentence-level translation while ensuring the absence of sentence-level omissions and upholding translation quality. It surpasses the other three methods in terms of s-COMET, d-BLEU, and d-COMET metrics.

\subsection{The results for the Multilingual test sets}

In addition to conducting a detailed comparison of methods in both Chinese-to-English and English-to-Chinese translation directions, we extended our experiments to include German, French, and Japanese with English, covering a total of eight language pairs. For a comprehensive view of the scores, refer to the appendix. Table~\ref{tab:Results_mul_lang} presents the average metrics for translations between other languages and English, as well as English and other languages, showing consistent trends with Table~\ref{tab:Results_two_test}. The Sentence method yields translations with lower d-ppl and LCTR-$1_f$, suggesting that its fluency and consistency are inadequate for document translation. While the Doc2Doc method exhibits good fluency and consistency, it fails to meet quality standards. The Doc2Sent method demonstrates higher COMET and BLEU scores, yet its d-ppl aligns with the Sentence method, indicating a trade-off between sentence fluency and quality/consistency that does not fully resolve the challenges inherent in document translation. In contrast, the Doc-Guided Sent2Sent++ method achieves favorable results across all metrics, maintaining quality and consistency comparable to or exceeding the Doc2Sent method, and matching the fluency of the Doc2Doc method.

\section{Ablation Study}

\subsection{Why our Doc-Guided Memory only use summary information?}

In extended experiments, we compared the method that incorporates both summaries and term information in Memory following \citet{DBLP:journals/corr/abs-2405-11804} with our approach—the Doc-Guided Sent2Sent++ Agent. Our method, by capturing the full-text summary, has already captured most of the terminology and stylistic nuances.

\begin{table}[!h]
\centering
% \footnotesize
% \begin{tabularx}{0.48\textwidth}{Xccc}
\begin{tabularx}{0.48\textwidth}{Xcc}
\hline
& Summary & Summary + Term \\
\hline
s-comet & 84.11 & 84.15  \\
d-bleu & 35.3 & 35.6  \\
d-comet & 6.66 & 6.65  \\
LCTR-$1_f$ & 92.12 & 92.18  \\
d-ppl & 16.18 & 16.16  \\
\hline
\end{tabularx}
\caption{\centering Summary v.s. Term}\label{tab:compare1}
\end{table}

As observed from the Table~\ref{tab:compare1}, the results show that this approach is practically equivalent to the method of real-time updating and addition of term information. Furthermore, this method significantly reduces the number of calls to large models and the length of context, thereby enhancing system efficiency.

\subsection{Do we need more context in the forced decoding of Sent2Sent++?}

In the decoding process of our Sent2Sent++, we only used the translation of the previous sentence for forced decoding. What if we use the translations of the first two sentences or more? 

\begin{table}[!h]
\centering
% \small
\begin{tabularx}{0.48\textwidth}{Xccc}
\hline
$n$ & 1 & 2 & 3 \\
\hline
s-comet & 84.11 & 84.13 & 83.95 \\
d-bleu & 35.3 & 34.8 & 35.4 \\
d-comet & 6.66 & 6.67 & 6.58 \\
LCTR-$1_f$ & 92.12 & 92.14 & 92.21 \\
d-ppl & 16.18 & 16.21 & 16.07 \\
\hline
\end{tabularx}
\caption{\centering $n$ previous context in Sent2Sent++}\label{tab:compare2}
\end{table}

As shown in Table \ref{tab:compare2}, we found no significant differences in various metrics as the number of preceding sentences increased. Therefore, considering efficiency, we only used the previous sentence.

\section{Related Work}

Many works have been published on the topic of document-level NMT. The widely used baseline approach consists of simply concatenating a few adjacent sentences and feeding this as an input to the MT system, without modifying the system architecture in any way \cite{DBLP:conf/discomt/TiedemannS17, DBLP:conf/naacl/BawdenSBH18, DBLP:conf/eamt/AgrawalTN18, DBLP:conf/iwslt/NguyenMC21}. Also, several modifications to this baseline concatenation approach have been proposed. \cite{DBLP:conf/acl/MaZZ20} introduce segment embeddings and also partially constrain the attention to the tokens of the current sentence. \cite{DBLP:conf/emnlp/ZhangCGF20} propose to calculate the self-attention both on the sentence- and on the document-level and then combine the two representations. \cite{DBLP:conf/acl/FernandesYNM20} and \cite{DBLP:conf/dasfaa/YangYMYGHZZLW23} both mask out tokens in the current sentence to increase context utilization while \cite{DBLP:conf/coling/LeiRX22} remove tokens from the context if they are not attended.

Beyond simple concatenation methods, there exist alternative document-level Neural Machine Translation (NMT) approaches. Sequential Decoding (SD) employs the previously generated target sentences as context to produce translations on a sentence-by-sentence basis. In contrast, our Sent2Sent++ method capitalizes on the rich information from the Doc-Guided Memory, necessitating only the information from one preceding sentence for each decoding step. 

\section{Conclusion}

In this paper, we have presented Doc-Guided Sent2Sent++, an innovative approach to Document-level Machine Translation (DocMT) that addresses the challenges of sentence omissions and document fluency. Our Agent, Sent2Sent++, employs an incremental sentence-level forced decoding strategy, which ensures the translation of every sentence while enhancing the fluency of adjacent sentences. Furthermore, we introduced the Doc-Guided Memory mechanism, which leverages abstract bilingual information as key contextual cues. This approach stands out from Doc2Doc and Doc2Sent methods by maintaining both the completeness and fluency of document translations. Our experiments across multiple languages and domains have demonstrated that Sent2Sent++ significantly outperforms existing methods in terms of s-COMET, d-COMET, LTCR-$1_f$, and perplexity (ppl), which are critical metrics for evaluating the quality of machine translation. We believe that our approach will pave the way for future research and development in this area, leading to even more sophisticated and accurate translation models.

\section{Limitation}

The Doc-Guided Sent2Sent++ Agent presented in this paper has demonstrated satisfactory results with respect to quality, consistency, and fluency in discourse translation. Although our experiments compared this method with approaches such as Doc2Doc and Doc2Sent, a direct comparison with commercial models, including GPT-4o, was not feasible. Owing to the decoding strategy of Sent2Sent++, the issue of multiple calls to large models persists. Future work will investigate the potential to harness the advanced capabilities of large models to perform forced decoding of multiple sentences simultaneously, with the goal of further enhancing translation efficiency without compromising quality.

\bibliography{custom}

\begin{thebibliography}{30}
\providecommand{\natexlab}[1]{#1}

\bibitem[{Agrawal et~al.(2018)Agrawal, Turchi, and Negri}]{DBLP:conf/eamt/AgrawalTN18}
Ruchit Agrawal, Marco Turchi, and Matteo Negri. 2018.
\newblock \href {https://aclanthology.org/2018.eamt-main.1} {Contextual handling in neural machine translation: Look behind, ahead and on both sides}.
\newblock In \emph{Proceedings of the 21st Annual Conference of the European Association for Machine Translation, {EAMT} 2018, Alicante, Spain, May 28-30, 2018}, pages 31--40.

\bibitem[{Bai et~al.(2023)Bai, Bai, Chu, Cui, Dang, Deng, Fan, Ge, Han, Huang, Hui, Ji, Li, Lin, Lin, Liu, Liu, Lu, Lu, Ma, Men, Ren, Ren, Tan, Tan, Tu, Wang, Wang, Wang, Wu, Xu, Xu, Yang, Yang, Yang, Yang, Yao, Yu, Yuan, Yuan, Zhang, Zhang, Zhang, Zhang, Zhou, Zhou, Zhou, and Zhu}]{DBLP:journals/corr/abs-2309-16609}
Jinze Bai, Shuai Bai, Yunfei Chu, Zeyu Cui, Kai Dang, Xiaodong Deng, Yang Fan, Wenbin Ge, Yu~Han, Fei Huang, Binyuan Hui, Luo Ji, Mei Li, Junyang Lin, Runji Lin, Dayiheng Liu, Gao Liu, Chengqiang Lu, Keming Lu, Jianxin Ma, Rui Men, Xingzhang Ren, Xuancheng Ren, Chuanqi Tan, Sinan Tan, Jianhong Tu, Peng Wang, Shijie Wang, Wei Wang, Shengguang Wu, Benfeng Xu, Jin Xu, An~Yang, Hao Yang, Jian Yang, Shusheng Yang, Yang Yao, Bowen Yu, Hongyi Yuan, Zheng Yuan, Jianwei Zhang, Xingxuan Zhang, Yichang Zhang, Zhenru Zhang, Chang Zhou, Jingren Zhou, Xiaohuan Zhou, and Tianhang Zhu. 2023.
\newblock \href {https://doi.org/10.48550/ARXIV.2309.16609} {Qwen technical report}.
\newblock \emph{CoRR}, abs/2309.16609.

\bibitem[{Bawden et~al.(2018)Bawden, Sennrich, Birch, and Haddow}]{DBLP:conf/naacl/BawdenSBH18}
Rachel Bawden, Rico Sennrich, Alexandra Birch, and Barry Haddow. 2018.
\newblock \href {https://doi.org/10.18653/V1/N18-1118} {Evaluating discourse phenomena in neural machine translation}.
\newblock In \emph{Proceedings of the 2018 Conference of the North American Chapter of the Association for Computational Linguistics: Human Language Technologies, {NAACL-HLT} 2018, New Orleans, Louisiana, USA, June 1-6, 2018, Volume 1 (Long Papers)}, pages 1304--1313. Association for Computational Linguistics.

\bibitem[{Cettolo et~al.(2017)Cettolo, Federico, Bentivogli, Niehues, St{\"{u}}ker, Sudoh, Yoshino, and Federmann}]{DBLP:conf/iwslt/CettoloFBNSSYF17}
Mauro Cettolo, Marcello Federico, Luisa Bentivogli, Jan Niehues, Sebastian St{\"{u}}ker, Katsuhito Sudoh, Koichiro Yoshino, and Christian Federmann. 2017.
\newblock \href {https://aclanthology.org/2017.iwslt-1.1} {Overview of the {IWSLT} 2017 evaluation campaign}.
\newblock In \emph{Proceedings of the 14th International Conference on Spoken Language Translation, {IWSLT} 2017, Tokyo, Japan, December 14-15, 2017}, pages 2--14. International Workshop on Spoken Language Translation.

\bibitem[{Cui et~al.(2024)Cui, Du, Zhu, and Xiong}]{DBLP:conf/acl/CuiDZX24}
Menglong Cui, Jiangcun Du, Shaolin Zhu, and Deyi Xiong. 2024.
\newblock \href {https://doi.org/10.18653/V1/2024.FINDINGS-ACL.646} {Efficiently exploring large language models for document-level machine translation with in-context learning}.
\newblock In \emph{Findings of the Association for Computational Linguistics, {ACL} 2024, Bangkok, Thailand and virtual meeting, August 11-16, 2024}, pages 10885--10897. Association for Computational Linguistics.

\bibitem[{Fernandes et~al.(2021)Fernandes, Yin, Neubig, and Martins}]{DBLP:conf/acl/FernandesYNM20}
Patrick Fernandes, Kayo Yin, Graham Neubig, and Andr{\'{e}} F.~T. Martins. 2021.
\newblock \href {https://doi.org/10.18653/V1/2021.ACL-LONG.505} {Measuring and increasing context usage in context-aware machine translation}.
\newblock In \emph{Proceedings of the 59th Annual Meeting of the Association for Computational Linguistics and the 11th International Joint Conference on Natural Language Processing, {ACL/IJCNLP} 2021, (Volume 1: Long Papers), Virtual Event, August 1-6, 2021}, pages 6467--6478. Association for Computational Linguistics.

\bibitem[{Karpinska and Iyyer(2023)}]{DBLP:conf/wmt/KarpinskaI23}
Marzena Karpinska and Mohit Iyyer. 2023.
\newblock \href {https://doi.org/10.18653/V1/2023.WMT-1.41} {Large language models effectively leverage document-level context for literary translation, but critical errors persist}.
\newblock In \emph{Proceedings of the Eighth Conference on Machine Translation, {WMT} 2023, Singapore, December 6-7, 2023}, pages 419--451. Association for Computational Linguistics.

\bibitem[{Kim et~al.(2019)Kim, Tran, and Ney}]{DBLP:conf/discomt/KimTN19}
Yunsu Kim, Duc~Thanh Tran, and Hermann Ney. 2019.
\newblock \href {https://doi.org/10.18653/V1/D19-6503} {When and why is document-level context useful in neural machine translation?}
\newblock In \emph{Proceedings of the Fourth Workshop on Discourse in Machine Translation, DiscoMT@EMNLP 2019, Hong Kong, China, November 3, 2019}, pages 24--34. Association for Computational Linguistics.

\bibitem[{Lei et~al.(2022)Lei, Ren, and Xiong}]{DBLP:conf/coling/LeiRX22}
Yikun Lei, Yuqi Ren, and Deyi Xiong. 2022.
\newblock \href {https://aclanthology.org/2022.coling-1.462} {Codonmt: Modeling cohesion devices for document-level neural machine translation}.
\newblock In \emph{Proceedings of the 29th International Conference on Computational Linguistics, {COLING} 2022, Gyeongju, Republic of Korea, October 12-17, 2022}, pages 5205--5216. International Committee on Computational Linguistics.

\bibitem[{Luo et~al.(2024)Luo, Guo, Wei, Shang, Li, Wu, Rao, Li, Yang, and Yang}]{DBLP:conf/wmt/LuoGWSLWRLYY24}
Yuanchang Luo, Jiaxin Guo, Daimeng Wei, Hengchao Shang, Zongyao Li, Zhanglin Wu, Zhiqiang Rao, Shaojun Li, Jinlong Yang, and Hao Yang. 2024.
\newblock \href {https://aclanthology.org/2024.wmt-1.97} {Context-aware and style-related incremental decoding framework for discourse-level literary translation}.
\newblock In \emph{Proceedings of the Ninth Conference on Machine Translation, {WMT} 2024, Miami, FL, USA, November 15-16, 2024}, pages 973--979. Association for Computational Linguistics.

\bibitem[{Lyu et~al.(2024)Lyu, Du, Xu, Duan, Wu, Lynn, Aji, Wong, and Wang}]{DBLP:conf/coling/LyuD0DWLAWW24}
Chenyang Lyu, Zefeng Du, Jitao Xu, Yitao Duan, Minghao Wu, Teresa Lynn, Alham~Fikri Aji, Derek~F. Wong, and Longyue Wang. 2024.
\newblock \href {https://aclanthology.org/2024.lrec-main.120} {A paradigm shift: The future of machine translation lies with large language models}.
\newblock In \emph{Proceedings of the 2024 Joint International Conference on Computational Linguistics, Language Resources and Evaluation, {LREC/COLING} 2024, 20-25 May, 2024, Torino, Italy}, pages 1339--1352. {ELRA} and {ICCL}.

\bibitem[{Lyu et~al.(2021)Lyu, Li, Gong, and Zhang}]{DBLP:conf/emnlp/LyuLGZ21}
Xinglin Lyu, Junhui Li, Zhengxian Gong, and Min Zhang. 2021.
\newblock \href {https://doi.org/10.18653/V1/2021.EMNLP-MAIN.262} {Encouraging lexical translation consistency for document-level neural machine translation}.
\newblock In \emph{Proceedings of the 2021 Conference on Empirical Methods in Natural Language Processing, {EMNLP} 2021, Virtual Event / Punta Cana, Dominican Republic, 7-11 November, 2021}, pages 3265--3277. Association for Computational Linguistics.

\bibitem[{Ma et~al.(2020)Ma, Zhang, and Zhou}]{DBLP:conf/acl/MaZZ20}
Shuming Ma, Dongdong Zhang, and Ming Zhou. 2020.
\newblock \href {https://doi.org/10.18653/V1/2020.ACL-MAIN.321} {A simple and effective unified encoder for document-level machine translation}.
\newblock In \emph{Proceedings of the 58th Annual Meeting of the Association for Computational Linguistics, {ACL} 2020, Online, July 5-10, 2020}, pages 3505--3511. Association for Computational Linguistics.

\bibitem[{Maruf et~al.(2022)Maruf, Saleh, and Haffari}]{DBLP:journals/csur/MarufSH21}
Sameen Maruf, Fahimeh Saleh, and Gholamreza Haffari. 2022.
\newblock \href {https://doi.org/10.1145/3441691} {A survey on document-level neural machine translation: Methods and evaluation}.
\newblock \emph{{ACM} Comput. Surv.}, 54(2):45:1--45:36.

\bibitem[{Nguyen et~al.(2021)Nguyen, Murray, and Chiang}]{DBLP:conf/iwslt/NguyenMC21}
Toan~Q. Nguyen, Kenton Murray, and David Chiang. 2021.
\newblock \href {https://doi.org/10.18653/V1/2021.IWSLT-1.33} {Data augmentation by concatenation for low-resource translation: {A} mystery and a solution}.
\newblock In \emph{Proceedings of the 18th International Conference on Spoken Language Translation, {IWSLT} 2021, Bangkok, Thailand (online), August 5-6, 2021}, pages 287--293. Association for Computational Linguistics.

\bibitem[{OpenAI(2023)}]{DBLP:journals/corr/abs-2303-08774}
OpenAI. 2023.
\newblock \href {https://doi.org/10.48550/ARXIV.2303.08774} {{GPT-4} technical report}.
\newblock \emph{CoRR}, abs/2303.08774.

\bibitem[{Tan et~al.(2021)Tan, Zhang, and Zhou}]{DBLP:conf/emnlp/TanZZ21}
Xin Tan, Longyin Zhang, and Guodong Zhou. 2021.
\newblock \href {https://doi.org/10.18653/V1/2021.EMNLP-MAIN.197} {Coupling context modeling with zero pronoun recovering for document-level natural language generation}.
\newblock In \emph{Proceedings of the 2021 Conference on Empirical Methods in Natural Language Processing, {EMNLP} 2021, Virtual Event / Punta Cana, Dominican Republic, 7-11 November, 2021}, pages 2530--2540. Association for Computational Linguistics.

\bibitem[{Tiedemann and Scherrer(2017)}]{DBLP:conf/discomt/TiedemannS17}
J{\"{o}}rg Tiedemann and Yves Scherrer. 2017.
\newblock \href {https://doi.org/10.18653/V1/W17-4811} {Neural machine translation with extended context}.
\newblock In \emph{Proceedings of the Third Workshop on Discourse in Machine Translation, DiscoMT@EMNLP 2017, Copenhagen, Denmark, September 8, 2017}, pages 82--92. Association for Computational Linguistics.

\bibitem[{Touvron et~al.(2023{\natexlab{a}})Touvron, Lavril, Izacard, Martinet, Lachaux, Lacroix, Rozi{\`{e}}re, Goyal, Hambro, Azhar, Rodriguez, Joulin, Grave, and Lample}]{DBLP:journals/corr/abs-2302-13971}
Hugo Touvron, Thibaut Lavril, Gautier Izacard, Xavier Martinet, Marie{-}Anne Lachaux, Timoth{\'{e}}e Lacroix, Baptiste Rozi{\`{e}}re, Naman Goyal, Eric Hambro, Faisal Azhar, Aur{\'{e}}lien Rodriguez, Armand Joulin, Edouard Grave, and Guillaume Lample. 2023{\natexlab{a}}.
\newblock \href {https://doi.org/10.48550/ARXIV.2302.13971} {Llama: Open and efficient foundation language models}.
\newblock \emph{CoRR}, abs/2302.13971.

\bibitem[{Touvron et~al.(2023{\natexlab{b}})Touvron, Martin, Stone, Albert, Almahairi, Babaei, Bashlykov, Batra, Bhargava, Bhosale, Bikel, Blecher, Canton{-}Ferrer, Chen, Cucurull, Esiobu, Fernandes, Fu, Fu, Fuller, Gao, Goswami, Goyal, Hartshorn, Hosseini, Hou, Inan, Kardas, Kerkez, Khabsa, Kloumann, Korenev, Koura, Lachaux, Lavril, Lee, Liskovich, Lu, Mao, Martinet, Mihaylov, Mishra, Molybog, Nie, Poulton, Reizenstein, Rungta, Saladi, Schelten, Silva, Smith, Subramanian, Tan, Tang, Taylor, Williams, Kuan, Xu, Yan, Zarov, Zhang, Fan, Kambadur, Narang, Rodriguez, Stojnic, Edunov, and Scialom}]{DBLP:journals/corr/abs-2307-09288}
Hugo Touvron, Louis Martin, Kevin Stone, Peter Albert, Amjad Almahairi, Yasmine Babaei, Nikolay Bashlykov, Soumya Batra, Prajjwal Bhargava, Shruti Bhosale, Dan Bikel, Lukas Blecher, Cristian Canton{-}Ferrer, Moya Chen, Guillem Cucurull, David Esiobu, Jude Fernandes, Jeremy Fu, Wenyin Fu, Brian Fuller, Cynthia Gao, Vedanuj Goswami, Naman Goyal, Anthony Hartshorn, Saghar Hosseini, Rui Hou, Hakan Inan, Marcin Kardas, Viktor Kerkez, Madian Khabsa, Isabel Kloumann, Artem Korenev, Punit~Singh Koura, Marie{-}Anne Lachaux, Thibaut Lavril, Jenya Lee, Diana Liskovich, Yinghai Lu, Yuning Mao, Xavier Martinet, Todor Mihaylov, Pushkar Mishra, Igor Molybog, Yixin Nie, Andrew Poulton, Jeremy Reizenstein, Rashi Rungta, Kalyan Saladi, Alan Schelten, Ruan Silva, Eric~Michael Smith, Ranjan Subramanian, Xiaoqing~Ellen Tan, Binh Tang, Ross Taylor, Adina Williams, Jian~Xiang Kuan, Puxin Xu, Zheng Yan, Iliyan Zarov, Yuchen Zhang, Angela Fan, Melanie Kambadur, Sharan Narang, Aur{\'{e}}lien Rodriguez, Robert Stojnic, Sergey Edunov,
  and Thomas Scialom. 2023{\natexlab{b}}.
\newblock \href {https://doi.org/10.48550/ARXIV.2307.09288} {Llama 2: Open foundation and fine-tuned chat models}.
\newblock \emph{CoRR}, abs/2307.09288.

\bibitem[{Vernikos et~al.(2022)Vernikos, Thompson, Mathur, and Federico}]{DBLP:conf/wmt/VernikosTMF22}
Giorgos Vernikos, Brian Thompson, Prashant Mathur, and Marcello Federico. 2022.
\newblock \href {https://aclanthology.org/2022.wmt-1.6} {Embarrassingly easy document-level {MT} metrics: How to convert any pretrained metric into a document-level metric}.
\newblock In \emph{Proceedings of the Seventh Conference on Machine Translation, {WMT} 2022, Abu Dhabi, United Arab Emirates (Hybrid), December 7-8, 2022}, pages 118--128. Association for Computational Linguistics.

\bibitem[{Wang et~al.(2024{\natexlab{a}})Wang, Liu, Lyu, Jiao, Wang, Xu, Tu, Gu, Chen, Wu, Zhou, Koehn, Way, and Yuan}]{DBLP:conf/wmt/WangLLJWXTGCWZK24}
Longyue Wang, Siyou Liu, Chenyang Lyu, Wenxiang Jiao, Xing Wang, Jiahao Xu, Zhaopeng Tu, Yan Gu, Weiyu Chen, Minghao Wu, Liting Zhou, Philipp Koehn, Andy Way, and Yulin Yuan. 2024{\natexlab{a}}.
\newblock \href {https://aclanthology.org/2024.wmt-1.58} {Findings of the {WMT} 2024 shared task on discourse-level literary translation}.
\newblock In \emph{Proceedings of the Ninth Conference on Machine Translation, {WMT} 2024, Miami, FL, USA, November 15-16, 2024}, pages 699--700. Association for Computational Linguistics.

\bibitem[{Wang et~al.(2017)Wang, Tu, Way, and Liu}]{DBLP:conf/emnlp/WangTWL17}
Longyue Wang, Zhaopeng Tu, Andy Way, and Qun Liu. 2017.
\newblock \href {https://doi.org/10.18653/V1/D17-1301} {Exploiting cross-sentence context for neural machine translation}.
\newblock In \emph{Proceedings of the 2017 Conference on Empirical Methods in Natural Language Processing, {EMNLP} 2017, Copenhagen, Denmark, September 9-11, 2017}, pages 2826--2831. Association for Computational Linguistics.

\bibitem[{Wang et~al.(2024{\natexlab{b}})Wang, Zeng, Liu, Wong, Meng, Zhou, and Zhang}]{DBLP:journals/corr/abs-2410-08143}
Yutong Wang, Jiali Zeng, Xuebo Liu, Derek~F. Wong, Fandong Meng, Jie Zhou, and Min Zhang. 2024{\natexlab{b}}.
\newblock \href {https://doi.org/10.48550/ARXIV.2410.08143} {Delta: An online document-level translation agent based on multi-level memory}.
\newblock \emph{CoRR}, abs/2410.08143.

\bibitem[{Wu et~al.(2024{\natexlab{a}})Wu, Vu, Qu, Foster, and Haffari}]{DBLP:journals/corr/abs-2401-06468}
Minghao Wu, Thuy{-}Trang Vu, Lizhen Qu, George~F. Foster, and Gholamreza Haffari. 2024{\natexlab{a}}.
\newblock \href {https://doi.org/10.48550/ARXIV.2401.06468} {Adapting large language models for document-level machine translation}.
\newblock \emph{CoRR}, abs/2401.06468.

\bibitem[{Wu et~al.(2024{\natexlab{b}})Wu, Yuan, Haffari, and Wang}]{DBLP:journals/corr/abs-2405-11804}
Minghao Wu, Yulin Yuan, Gholamreza Haffari, and Longyue Wang. 2024{\natexlab{b}}.
\newblock \href {https://doi.org/10.48550/ARXIV.2405.11804} {(perhaps) beyond human translation: Harnessing multi-agent collaboration for translating ultra-long literary texts}.
\newblock \emph{CoRR}, abs/2405.11804.

\bibitem[{Wu and Hu(2023)}]{DBLP:conf/wmt/WuH23}
Yangjian Wu and Gang Hu. 2023.
\newblock \href {https://doi.org/10.18653/V1/2023.WMT-1.15} {Exploring prompt engineering with {GPT} language models for document-level machine translation: Insights and findings}.
\newblock In \emph{Proceedings of the Eighth Conference on Machine Translation, {WMT} 2023, Singapore, December 6-7, 2023}, pages 166--169. Association for Computational Linguistics.

\bibitem[{Yang et~al.(2024)Yang, Yang, Hui, Zheng, Yu, Zhou, Li, Li, Liu, Huang, Dong, Wei, Lin, Tang, Wang, Yang, Tu, Zhang, Ma, Yang, Xu, Zhou, Bai, He, Lin, Dang, Lu, Chen, Yang, Li, Xue, Ni, Zhang, Wang, Peng, Men, Gao, Lin, Wang, Bai, Tan, Zhu, Li, Liu, Ge, Deng, Zhou, Ren, Zhang, Wei, Ren, Liu, Fan, Yao, Zhang, Wan, Chu, Liu, Cui, Zhang, Guo, and Fan}]{DBLP:journals/corr/abs-2407-10671}
An~Yang, Baosong Yang, Binyuan Hui, Bo~Zheng, Bowen Yu, Chang Zhou, Chengpeng Li, Chengyuan Li, Dayiheng Liu, Fei Huang, Guanting Dong, Haoran Wei, Huan Lin, Jialong Tang, Jialin Wang, Jian Yang, Jianhong Tu, Jianwei Zhang, Jianxin Ma, Jianxin Yang, Jin Xu, Jingren Zhou, Jinze Bai, Jinzheng He, Junyang Lin, Kai Dang, Keming Lu, Keqin Chen, Kexin Yang, Mei Li, Mingfeng Xue, Na~Ni, Pei Zhang, Peng Wang, Ru~Peng, Rui Men, Ruize Gao, Runji Lin, Shijie Wang, Shuai Bai, Sinan Tan, Tianhang Zhu, Tianhao Li, Tianyu Liu, Wenbin Ge, Xiaodong Deng, Xiaohuan Zhou, Xingzhang Ren, Xinyu Zhang, Xipin Wei, Xuancheng Ren, Xuejing Liu, Yang Fan, Yang Yao, Yichang Zhang, Yu~Wan, Yunfei Chu, Yuqiong Liu, Zeyu Cui, Zhenru Zhang, Zhifang Guo, and Zhihao Fan. 2024.
\newblock \href {https://doi.org/10.48550/ARXIV.2407.10671} {Qwen2 technical report}.
\newblock \emph{CoRR}, abs/2407.10671.

\bibitem[{Yang et~al.(2023)Yang, Yin, Ma, Yang, Guo, Huang, Zhang, Zeng, Li, and Wei}]{DBLP:conf/dasfaa/YangYMYGHZZLW23}
Jian Yang, Yuwei Yin, Shuming Ma, Liqun Yang, Hongcheng Guo, Haoyang Huang, Dongdong Zhang, Yutao Zeng, Zhoujun Li, and Furu Wei. 2023.
\newblock \href {https://doi.org/10.1007/978-3-031-30675-4\_34} {Hanoit: Enhancing context-aware translation via selective context}.
\newblock In \emph{Database Systems for Advanced Applications - 28th International Conference, {DASFAA} 2023, Tianjin, China, April 17-20, 2023, Proceedings, Part {III}}, volume 13945 of \emph{Lecture Notes in Computer Science}, pages 471--486. Springer.

\bibitem[{Zhang et~al.(2020)Zhang, Chen, Ge, and Fan}]{DBLP:conf/emnlp/ZhangCGF20}
Pei Zhang, Boxing Chen, Niyu Ge, and Kai Fan. 2020.
\newblock \href {https://doi.org/10.18653/V1/2020.EMNLP-MAIN.81} {Long-short term masking transformer: {A} simple but effective baseline for document-level neural machine translation}.
\newblock In \emph{Proceedings of the 2020 Conference on Empirical Methods in Natural Language Processing, {EMNLP} 2020, Online, November 16-20, 2020}, pages 1081--1087. Association for Computational Linguistics.

\end{thebibliography}

\clearpage
\onecolumn
\appendix

\section{Detailed Results for Multilingual Tests}
\label{sec:appendix}

\begin{table*}[th]
\centering
% \footnotesize
\begin{tabularx}{\textwidth}{X|Xccccc}
\hline
\textbf{lang-pair} & \textbf{method} & \textbf{s-comet} & \textbf{d-bleu}  & \textbf{d-comet} & \textbf{LCTR-$1_f$} & \textbf{d-ppl} \\
\hline
\multirow{6}*{zh2en} & Sentence & 83.22
  & 26.6 & 7.14 & 87.69 & 11.28 \\
 & Doc2Doc & -  & 23.25 & - &  93.05 & 9.19 \\
 & Doc2Sent & 83.34  & 26.8 & 7.15 &  92.84  & 11.1 \\
 & \ \ \ \ -\ Memory & 83.27  & 26.3 & 7.12 &  87.96  & 11.3 \\
\cline{2-7}
 & Sent2Sent++ & 83.34  & 28.3 & 7.25 &  92.85  & 10.49 \\
 & \ \ \ \ -\ Memory & 83.39  & 27.9 & 7.27 &  87.47  & 10.32 \\

\hline
\multirow{6}*{de2en} & Sentence & 85.92
  & 33.25 & 7.17 & 93.12 & 11.63 \\
 & Doc2Doc & -  & 30.99 & - &  98.05 & 9.62 \\
 & Doc2Sent & 86.17  & 33.27 & 7.18 &  98.13  & 11.44 \\
 & \ \ \ \ -\ Memory & 86.04  & 33.12 & 7.16 &  94.23  & 11.54 \\
\cline{2-7}
 & Sent2Sent++ & 86.27  & 34.76 & 7.25 &  98.64  & 10.89 \\
 & \ \ \ \ -\ Memory & 86.19  & 34.39 & 7.22 &  94.32  & 10.93 \\
\hline

\hline
\multirow{6}*{fr2en} & Sentence & 87.32
  & 42.78 & 6.71 & 89.21 & 11.6 \\
 & Doc2Doc & -  & 38.46 & - &  91.68 & 10.36 \\
 & Doc2Sent & 87.3  & 42.28 & 6.73 &  93.02  & 11.56 \\
 & \ \ \ \ -\ Memory & 87.15  & 42.02 & 6.69 &  90.05  & 11.62 \\
\cline{2-7}
 & Sent2Sent++ & 87.51  & 43.66 & 6.76 &  93.16  & 11.26 \\
 & \ \ \ \ -\ Memory & 87.36  & 43.38 & 6.71 &  90.65  & 11.34 \\

\hline
\multirow{6}*{ja2en} & Sentence & 80.57
  & 18.36 & 6.84 & 54.36 & 12.16 \\
 & Doc2Doc & -  & 14.34 & - &  68.72 & 6.03 \\
 & Doc2Sent & 80.75  & 18.59 & 6.91 &  71.14  & 11.37 \\
 & \ \ \ \ -\ Memory & 80.79  & 18.62 & 6.92 &  56.07  & 11.42 \\
\cline{2-7}
 & Sent2Sent++ & 80.51  & 19.24 & 7.09 &  72.61  & 10.14 \\
 & \ \ \ \ -\ Memory & 79.97  & 19.31 & 7.11 &  56.29  & 10.18 \\

\hline

\end{tabularx}
\caption{\centering Results for xx2en directions in the IWSLT2017 test set}\label{tab:Results_mul_langxx}
\end{table*}

\begin{table*}[th]
\centering
% \footnotesize
\begin{tabularx}{\textwidth}{X|Xccccc}
\hline
\textbf{lang-pair} & \textbf{method} & \textbf{s-comet} & \textbf{d-bleu}  & \textbf{d-comet} & \textbf{LCTR-$1_f$} & \textbf{d-ppl} \\
\hline
\multirow{4}*{en2zh} & Sentence & 82.9
  & 35.1 & 6.41 & 83.79 & 18.25 \\
 & Doc2Doc & -  & 36.58 & - &  90.65 & 16.69 \\
 & Doc2Sent & 83.42  & 35.7 & 6.44 &  92.14  & 17.53 \\
 & \ \ \ \ -\ Memory & 83.26  & 35.3 & 6.43 &  83.93  & 17.54 \\
 & Sent2Sent++ & 84.11  & 35.3 & 6.66 &  92.12  & 16.18 \\
 & \ \ \ \ -\ Memory & 83.94  & 34.9 & 6.56 &  83.95  & 16.23 \\

\hline
\multirow{4}*{en2de} & Sentence & 83.68
  & 28.21 & 6.51 & 82.01 & 7.7 \\
 & Doc2Doc & -  & 27.03 & - &  85.98 & 6.70 \\
 & Doc2Sent & 83.71  & 28.27 & 6.48 &  87.60  & 7.72 \\
 & \ \ \ \ -\ Memory & 83.69  & 28.12 & 6.46 &  82.01  & 7.78 \\
 & Sent2Sent++ & 83.96  & 29.15 & 6.62 &  88.93  & 7.4 \\
 & \ \ \ \ -\ Memory & 83.84  & 28.92 & 6.57 &  82.29  & 7.45 \\
\hline

\hline
\multirow{4}*{en2fr} & Sentence & 84.74
  & 39.26 & 5.98 & 81.92 & 6.62 \\
 & Doc2Doc & -  & 36.54 & - &  85.72 & 6.15 \\
 & Doc2Sent & 84.45  & 39.65 & 6.04 &  88.42  & 6.65 \\
 & \ \ \ \ -\ Memory & 84.29  & 39.42 & 5.98 &  82.09  & 6.7 \\
 & Sent2Sent++ & 84.78  & 39.8 & 6.06 &  88.99  & 6.47 \\
 & \ \ \ \ -\ Memory & 84.64  & 39.2 & 6.01 &  82.48  & 6.52 \\

\hline
\multirow{4}*{en2ja} & Sentence & 82.59
  & 15.89 & 6.71 & 54.19 & 11.11 \\
 & Doc2Doc & -  & 14.62 & - &  60.13 & 8.61 \\
 & Doc2Sent & 83.32  & 15.93 & 6.81 &  66.36  & 10.72 \\
 & \ \ \ \ -\ Memory & 83.29  & 14.87 & 6.72 &  54.34  & 10.81 \\
 & Sent2Sent++ & 82.26  & 15.08 & 6.88 & 67.42 & 8.71 \\
 & \ \ \ \ -\ Memory & 82.09  & 14.9 & 6.76 & 54.31 & 8.79 \\

\hline

\end{tabularx}
\caption{\centering Results for en2xx directions in the IWSLT2017 test set}\label{tab:Results_mul_langxxx}
\end{table*}

\end{document}